\pdfoutput=1

\documentclass[11pt]{article}

\usepackage[]{acl}
\usepackage{graphicx}
\usepackage{breqn}
\usepackage{bbold}
\usepackage{tikz}
\usepackage{multirow}
\usepackage{caption}
\usepackage{subcaption}
\usepackage{amsmath}
\usepackage{amssymb}
\usepackage{xurl}
\usepackage{times}
\usepackage{latexsym}
\usepackage{comment}
\usepackage{float}
\usepackage{chngpage}
\usepackage{microtype}
\usepackage{cuted}
\usepackage{booktabs}
\usepackage{algorithm}
\usepackage{algpseudocode}
\usepackage{tabulary}
\usepackage{array}
\usepackage{xcolor}

\usepackage[T1]{fontenc}
\usepackage[utf8]{inputenc}
\usepackage{microtype}
\title{Incorporating Attribution Importance for Improving Faithfulness Metrics}

\author{Zhixue Zhao \quad  Nikolaos Aletras \\
        Department of Computer Science, University of Sheffield \\
        United Kingdom \\
\texttt{\{zhixue.zhao, n.aletras\}@sheffield.ac.uk}}

\begin{document}
\maketitle

\begin{abstract}

Feature attribution methods (FAs) are popular approaches for providing insights into the model reasoning process of making predictions.
The more faithful a FA is, the more accurately it reflects which parts of the input are more important for the prediction.
Widely used faithfulness metrics, such as sufficiency and comprehensiveness use a hard erasure criterion, i.e. entirely removing or retaining the top most important tokens ranked by a given FA and observing the changes in predictive likelihood. 
However, this hard criterion ignores the importance of each individual token, treating them all equally for computing sufficiency and comprehensiveness. In this paper, we propose a simple yet effective soft erasure criterion. Instead of entirely removing or retaining tokens from the input, we randomly mask parts of the token vector representations proportionately to their FA importance. Extensive experiments across various natural language processing tasks and different FAs show that our soft-sufficiency and soft-comprehensiveness metrics consistently prefer more faithful explanations compared to hard sufficiency and comprehensiveness.\footnote{Our code: \url{https://github.com/casszhao/SoftFaith}}
\end{abstract}

\section{Introduction}
Feature attribution methods (FAs) are popular post-hoc explanation methods that are applied after model training to assign an importance score to each token in the input~\cite{kindermans2016investigating,pmlr-v70-sundararajan17a}. These scores indicate how much each token contributes to the model prediction. Typically, the top-k ranked tokens are then selected to form an explanation, i.e. rationale~\citep{deyoung2020eraser}. However, it is an important challenge to choose a FA for a natural language processing (NLP) task at hand~\cite{chalkidis-etal-2021-paragraph,fomicheva-etal-2022-translation} since there is no single FA that is consistently more faithful~\cite{atanasova-etal-2020-diagnostic}. 

To assess whether a rationale extracted with a given FA is faithful, i.e. actually reflects the true model reasoning~\cite{jacovi-goldberg-2020-towards}, various faithfulness metrics have been proposed~\cite{arras2017relevant,serrano-smith-2019-attention,jain-wallace-2019-attention,deyoung2020eraser}. Sufficiency and comprehensiveness~\cite{deyoung2020eraser}, also referred to as fidelity metrics~\cite{carton-etal-2020-evaluating}, are two widely used metrics which have been found to be effective in capturing rationale faithfulness~\cite{chrysostomou-aletras-2021-enjoy, chan-etal-2022-comparative}. Both metrics use a hard erasure criterion for perturbing the input by entirely removing (i.e. comprehensiveness) or retaining (i.e. sufficiency) the rationale to observe changes in predictive likelihood. 

\begin{figure}[!t]
\centering
\includegraphics[width=0.5\textwidth, trim={0cm 0cm 0cm 0cm}]{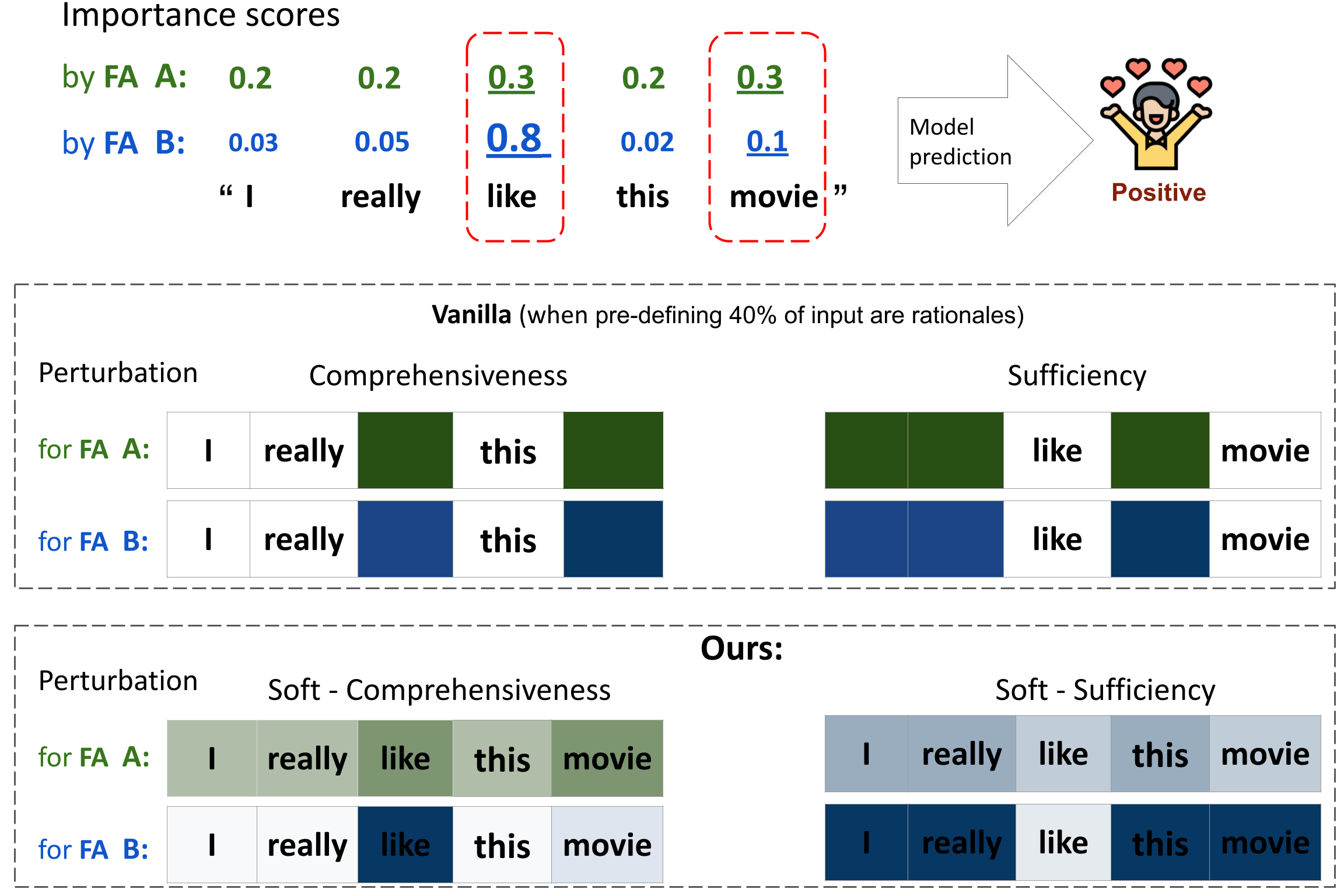}
\caption{Hard and soft erasure criteria for comprehensiveness and sufficiency for two toy feature attribution (FA) methods A and B. 
}\label{fig:idea}
\end{figure}

However, the hard erasure criterion ignores the different importance of each individual token, treating them all equally for computing sufficiency and comprehensiveness. Moreover, the hard-perturbed input is likely to fall out of the distribution the model is trained on, leading to inaccurate measurements of faithfulness~\cite{bastings-filippova-2020-elephant,yin-etal-2022-sensitivity,chrysostomou-aletras-2022-empirical,zhao-etal-2022-impact}. Figure~\ref{fig:idea} shows an example of two toy FAs, A and B, identifying the same top two tokens (``like'', ``movie'') as a rationale for the prediction. Still, each of them assigns different importance scores to the two tokens resulting into different rankings. According to the hard erasure criterion, comprehensiveness and sufficiency will assign the same faithfulness score to the two rationales extracted by the two FAs.

In this paper, we aim to improve sufficiency and comprehensiveness in capturing the faithfulness of a FA. We achieve this by replacing the hard token perturbation with a simple yet effective soft erasure criterion  (see Figure~\ref{fig:idea} for an intuitive example). Instead of entirely removing or retaining tokens from the input, we randomly mask parts of token vector representations proportionately to their FA importance.

Our main contributions are as follows:
\begin{itemize}
    
    \item We propose two new faithfulness metrics, soft-comprehensiveness and soft-sufficiency that rely on soft perturbations of the input. Our metrics are more robust to distribution shifts by avoiding entirely masking whole tokens; 
    \item We demonstrate that our metrics are consistently more effective in terms of preferring more faithful rather than unfaithful (i.e. random) FAs~\cite{chan-etal-2022-comparative}, compared to their ``hard'' counterparts across various NLP tasks and different FAs. 
    \item We advocate for evaluating the faithfulness of FAs by taking into account the entire input rather than manually pre-defining rationale lengths. 

\end{itemize}

\section{Related Work}

\subsection{Feature Attribution Methods}

A popular approach to assign token importance is by computing the gradients of the predictions with respect to the input~\citep{kindermans2016investigating,pmlr-v70-shrikumar17a,pmlr-v70-sundararajan17a}. A different approach is based on making perturbations in the input or individual neurons aiming to capture their impact on later neurons~\citep{zeiler2014visualizing}. In NLP, attention mechanism scores have been extensively used for assigning token importance~\cite{jain-wallace-2019-attention,serrano-smith-2019-attention,treviso-martins-2020-explanation, chrysostomou-aletras-2021-improving}. Finally, a  widely used group of FA methods is based on training simpler linear meta-models to assign token importance~\citep{ribeiro-etal-2016-trust}.

Given the large variety of approaches, it is often hard to choose an optimal FA for a given task. Previous work has demonstrated that different FAs generate inconsistent or conflicting explanations for the same model on the same input~\cite{atanasova-etal-2020-diagnostic, zhao-etal-2022-impact}.

\subsection{Measuring Faithfulness}
One standard approach to compare FAs and their rationales is faithfulness. 
A faithful model explanation is expected to accurately represent the true reasoning process of the model \citep{jacovi-goldberg-2020-towards}.

The majority of existing methods for quantitatively evaluating faithfulness is based on input perturbation \citep{nguyen-2018-comparing,deyoung2020eraser,ju-etal-2022-logic}. The main idea is to modify the input by entirely removing or retaining tokens according to their FA scores aiming to measure the difference in predictive likelihood . 

Commonly-used perturbation methods include comprehensiveness, i.e. removing the rationale from the input), and sufficiency, i.e. retaining only the rationale \citep{deyoung2020eraser}. 
Another common approach is to remove a number of tokens and observe the number of times the predicted label changes, i.e. Decision Flip \citep{serrano-smith-2019-attention}. 
On the other hand, Monotonicity incrementally adds more important tokens while Correlation between Importance and Output Probability (CORR) continuously removes the most important tokens \citep{arya2021one}. (In)fidelity perturbs the input by dropping a number of tokens in a decreasing order of attribution scores until the prediction changes~\cite{zafar-etal-2021-lack}.
Additionally, \citet{yin-etal-2022-sensitivity} proposed sensitivity and stability, which do not directly remove or keep tokens. Sensitivity adds noise to the entire rationale set aiming to find a minimum noise threshold for causing a prediction flip. Stability compares the predictions on semantically similar inputs. 

One limitation of the metrics above is that they ignore the relative importance of each individual token within the selected rationale, treating all of them equally. Despite the fact that some of them might take the FA ranking into account, the relative importance is still not considered. \citet{jacovi-goldberg-2020-towards} have emphasized that faithfulness should be evaluated on a ``grayscale'' rather than ``binary'' (i.e. faithful or not) manner. However, current perturbation-based metrics, such as comprehensiveness and sufficiency, do not reflect a ``grayscale'' fashion as tokens are entirely removed or retained (e.g. comprehensiveness, sufficiency), or the rationale is entirely perturbed as a whole (e.g. sensitivity).

\subsection{Evaluating Faithfulness Metrics}

Quantitatively measuring the faithfulness of model explanations is an open research problem with several recent efforts focusing on highlighting the main issues of current metrics~\cite{bastings-filippova-2020-elephant,ju-etal-2022-logic,yin-etal-2022-sensitivity} and comparing their effectiveness~\cite{chan-etal-2022-comparative}. 

A main challenge in comparing faithfulness metrics is that there is no access to ground truth, i.e. the true rationale for a model prediction~\citep{jacovi-goldberg-2020-towards,ye-etal-2021-connecting,lyu2022towards,ju-etal-2022-logic}. Additionally, \citet{ju-etal-2022-logic} argue that it is risky to design faithfulness metrics based on the assumption that a faithful FA will generate consistent or similar explanations for similar inputs and inconsistent explanations for adversarial inputs~\citep{alvarez2018robustness,sinha-etal-2021-perturbing,yin-etal-2022-sensitivity}. 

\citet{chan-etal-2022-comparative} introduced diagnosticity for comparing the effectiveness of faithfulness metrics. Diagnosticity measures the ability of a metric on separating random explanations (non-faithful) and non-random ones (faithful).  
They empirically showed that two perturbation metrics, sufficiency and comprehensiveness, are more `diagnostic', i.e. effective in choosing faithful rationales compared to other metrics.

Despite the fact that sufficiency and comprehensiveness are in general more effective, they suffer from an out-of-distribution issue~\cite{ancona2018towards,bastings-filippova-2020-elephant,yin-etal-2022-sensitivity}. More specifically, the hard perturbation (i.e. entirely removing or retaining tokens) creates a discretely corrupted version of the original input which might fall out of the distribution the model was trained on. It is unlikely that the model predictions over the corrupted input sentences share the same reasoning process with the original full sentences which might be misleading for uncovering the model's true reasoning mechanisms.

\section{Faithfulness Evaluation Metrics}\label{sec:faithfulness_metrics}

\subsection{Sufficiency and Comprehensiveness}

We begin by formally defining sufficiency and comprehensiveness~\citep{deyoung2020eraser}, and their corresponding normalized versions that allow for a fairer comparison across models and tasks proposed by \citet{carton-etal-2020-evaluating}.

\paragraph{Normalized Sufficiency (NS):} Sufficiency (S) aims to capture the difference in predictive likelihood between retaining only the rationale $p(\hat{y}|\mathcal{R})$ and the full text model $p(\hat{y}| \mathbf{X})$. We use the normalized version:
\begin{equation}
\label{equ:Norm_Suff}
\small
\begin{aligned}
    \text{S}(\mathbf{X}, \hat{y}, \mathcal{R}) = 1 - max(0, p(\hat{y}| \mathbf{X})- p (\hat{y}|\mathcal{R})) \\
    \text{NS}(\mathbf{X}, \hat{y}, \mathcal{R}) = \frac{\text{S}(\mathbf{X}, \hat{y}, \mathcal{R}) - \text{S}(\mathbf{X}, \hat{y}, 0)}{1 - \text{S}(\mathbf{X}, \hat{y}, 0)}
 \end{aligned}
\end{equation}

\noindent where $\text{S}(\mathbf{x}, \hat{y}, 0)$ is the sufficiency of a baseline input (zeroed out sequence) and $\hat{y}$ is the model predicted class using the full text $\mathbf{x}$ as input.

\paragraph{Normalized Comprehensiveness (NC):} Comprehensiveness (C) assesses how much information the rationale holds by measuring changes in predictive likelihoods when removing the rationale $p(\hat{y}|\mathbf{X}_{\backslash\mathcal{R}})$. The normalized version is defined as: 
\begin{equation}
\label{equ:Norm_Comp}
    \small
    \begin{aligned}
        \text{C}(\mathbf{X}, \hat{y}, \mathcal{R}) = max(0, p(\hat{y}| \mathbf{X})- p (\hat{y}|\mathbf{X}_{\backslash\mathcal{R}})) \\\\
        \text{NC}(\mathbf{X}, \hat{y}, \mathcal{R}) = \frac{\text{C}(\mathbf{X}, \hat{y}, \mathcal{R})}{1 - \text{S}(\mathbf{X}, \hat{y}, 0)}
    \end{aligned}
\end{equation}

\subsection{Soft Nomralized Sufficiency and Comprehensiveness}
Inspired by recent work that highlights the out-of-distribution issues of hard input perturbation~\cite{bastings-filippova-2020-elephant,yin-etal-2022-sensitivity, zhao-etal-2022-impact}, our goal is to induce to sufficiency and comprehensiveness the relative importance of all tokens determined by a given FA. For this purpose, we propose Soft Normalized Sufficiency (Soft-NS) and Soft Normalized Comprehensiveness (Soft-NC) that apply a soft-erasure criterion to perturb the input.

\paragraph{Soft Input Perturbation:}  

Given the vector representation of an input token, we aim to retain or remove vector elements proportionately to the token importance assigned by a FA by applying a Bernoulli distribution mask to the token embedding. 
Given a token vector $\mathbf{x}_i$ from the input $\mathbf{X}$ and its FA score $a_i$, we soft-perturb the input as follows:
\begin{align}
           \mathbf{x'_i} = \mathbf{x_i} \odot \mathbf{e_i},\: \mathbf{e_i}\sim \operatorname {Ber} (q)
           \label{eq:soft-perturbation}
\end{align}
\noindent where $Ber$ a Bernoulli distribution and $\mathbf{e}$ a binary mask vector of size $n$. $Ber$ is parameterized with probability $q$: 
\[
   q = 
\begin{cases}
    a,   & \text{if retaining elements}\\
    1-a, & \text{if removing elements}
\end{cases}
\]

\noindent We repeat the soft-perturbation for all token embeddings in the input to obtain $\mathbf{x'}$. Our approach is a special case of dropout~\citep{srivastava2014dropout} on the embedding level.

Following~\citet{lakshmi-narayan-etal-2019-exploration}, we have also tested two other approaches to soft perturbation in early-experimentation: (1) adding Gaussian noise to the embeddings; and (2) perturbing the attention scores, both in proportion to the FA scores. However, we found that dropout outperforms these two methods. Perhaps this is due to their sensitivity to hyperparameter tuning (e.g. standard deviation) which potentially contributes to their poor performance. Hence, we only conduct full experiments using dropout-based soft perturbation. Details on these alternative methods to perturb the input are included in Appendix \ref{app:alternative_soft_perturbation}.

\paragraph{Soft Normalized Sufficiency (Soft-NS):} 

The main assumption of Soft-NS is that the more important a token is, the larger number of embedding elements should be retained. On the other hand, if a token is not important most of its elements should be dropped. This way Soft-NS takes into account the complete ranking and importance scores of the FA while NS only keeps the top-k important tokens by ignoring their FA scores. We compute Soft-NS as follows:%

{\small
\begin{align}
\label{equ:Soft_Suff}
    \text{Soft-S}(\mathbf{X}, \hat{y}, \mathbf{X'}) = 1 - max(0, p(\hat{y}| \mathbf{X})- p (\hat{y}|\mathbf{X'})) \notag\\
    \notag\\
    \text{Soft-NS}(\mathbf{X}, \hat{y}, \mathbf{X'}) =
    \frac{\text{Soft-S}(\mathbf{X}, \hat{y}, \mathbf{X'}) - \text{S}(\mathbf{X}, \hat{y}, 0)}{1 - \text{S}(\mathbf{X}, \hat{y}, 0)}
\end{align}
}%
\noindent where $\mathbf{X'}$ is obtained by using $q=a_i$ in Eq.~\ref{eq:soft-perturbation} for each token vector $\mathbf{x'_i}$.

\paragraph{Soft Normalized Comprehensiveness (Soft-NC):} For Soft-NC, we assume that the more important a token is to the model prediction, the heavier the perturbation to its embedding should be. Soft-NS is computed as:

\begin{equation}
\label{equ:Soft_Comp}
\small
\begin{aligned}
    \text{Soft-C}(\mathbf{X}, \hat{y}, \mathbf{X'}) = max(0, p(\hat{y}| \mathbf{X})- p (\hat{y}|\mathbf{X'})) \\\\
    \text{Soft-NC}(\mathbf{X}, \hat{y}, \mathbf{X'}) = \frac{\text{Soft-C}(\mathbf{X}, \hat{y}, \mathbf{X'})}{1 - \text{S}(\mathbf{X}, \hat{y}, 0)}
\end{aligned}
\end{equation}
\noindent where $\mathbf{X'}$ is obtained by using $q=1-a_i$ in Eq.~\ref{eq:soft-perturbation} for each token vector $\mathbf{x'_i}$.


\section{Experimental Setup}

\subsection{Tasks}

Following related work on interpretability \cite{jain2020learning,chrysostomou2022flexible}, we experiment with the following datasets:
\begin{itemize}
    \item {\bf SST}: Binary sentiment classification into positive and negative classes \citep{socher-etal-2013-recursive}. 
    \item {\bf AG}: News articles categorized in Science, Sports, Business, and World topics \citep{del2005ranking}. 
    \item {\bf Evidence Inference (Ev.Inf.)}: Abstract-only biomedical articles describing randomized controlled trials. The task is to infer the relationship between a given intervention and comparator with respect to an outcome \citep{lehman-etal-2019-inferring}. 
    \item {\bf MultiRC (M.RC)}: A reading comprehension task with questions having multiple correct answers that should inferred from information from multiple sentences \citep{khashabi-etal-2018-looking}. Following \citet{deyoung2020eraser} and \citet{jain2020learning}, we convert this to a binary classification task where each rationale/question/answer triplet forms an instance and each candidate answer is labelled as True/False.
\end{itemize}

\subsection{Models}
Following \citet{jain2020learning}, we use BERT \citep{devlin2019bert} for SST and AG; SCIBERT \citep{beltagy-etal-2019-scibert} for EV.INF. and RoBERTa \citep{liu2019roberta} for M.RC. See App. \ref{app:model_hyperparameters} for hyperparameters. Dataset statistics and model prediction performance are shown in Table \ref{tab:data_statistics}.

\begin{table}[!t]
\resizebox{\columnwidth}{!}{%
\renewcommand*{\arraystretch}{1.1}
\begin{tabular}{@{}ccccc@{}}
\toprule
Dataset & Avg. Length & Classes & Size (Train/Dev/Test) & Avg. F1 \\ \midrule
SST & 18 & 2 & 6,920 / 872 / 1,821 & 90.4 $\pm$ 0.5 \\
AG & 36 & 4 & 102,000 / 18,000 / 7,600 & 93.6 $\pm$ 0.2 \\
Ev.Inf & 363 & 3 & 5,789 / 684 / 720 & 82.3 $\pm$ 2.2 \\
M.RC & 305 & 2 & 24,029 / 3,214 / 4,848 & 74.0 $\pm$ 2.5 \\ \bottomrule
\end{tabular}%
}
\caption{Dataset statistics and mode prediction performance (average over five runs)}
\label{tab:data_statistics}
\end{table}

\subsection{Feature Attribution Methods}
\label{sec:FAs}

We experiment with several popular feature attribution methods to compare faithfulness metrics. We do not focus on benchmarking various FAs but to improve faithfulness evaluation metrics.

\begin{itemize}
\item{\bf Attention ($\alpha$):} Token importance is computed using the corresponding normalized attention score \citep{jain2020learning}.

\item{\bf Scaled attention ($\alpha\nabla\alpha$):} Attention scores scaled by their corresponding gradients \citep{serrano-smith-2019-attention}.

\item{\bf InputXGrad ($x\nabla x$):} It attributes importance by multiplying the input with its gradient computed with respect to the predicted class \citep{kindermans2016investigating, atanasova-etal-2020-diagnostic}.

\item{\bf Integrated Gradients (IG):} This FA ranks input tokens by computing the integral of the gradients taken along a straight path from a baseline input (i.e. zero embedding vector) to the original input \citep{pmlr-v70-sundararajan17a}.

\item{\bf DeepLift (DL):} It computes token importance according to the difference between the activation of each neuron and a reference activation, i.e. zero embedding vector \citep{pmlr-v70-shrikumar17a}.
\end{itemize}

\subsection{Computing Faithfulness with Normalized Sufficiency and Comprehensiveness}

Following \citet{deyoung2020eraser}, we compute the Area Over the Perturbation Curve (AOPC) for normalized sufficiency (NS) and comprehensiveness (NC) across different rationale lengths. AOPC provides a better overall estimate of faithfulness~\cite{deyoung2020eraser}. We evaluate five different rationale ratios set to 1\%, 5\%, 10\%, 20\% and 50\%, similar to \citet{deyoung2020eraser} and \citet{chan-etal-2022-comparative}.

\subsection{Comparing the Diagnosticity of Faithfulness Metrics}

Comparing faithfulness metrics is a challenging task because there is no a priori ground truth rationales that can be used. 

\paragraph{Diagnosticity:} \citet{chan-etal-2022-comparative} proposed diagnosticity to measure the degree of a given faithfulness metric favors more faithful rationales over less faithful ones. The assumption behind this metric is that the importance scores assigned by a FA are highly likely to be more faithful than simply assigning random importance scores to tokens.  Given an explanation pair $(u, v)$, the diagnosticity is measured as the probability of $u$ being a more faithful explanation than $v$ given the same faithfulness metric $F$. $u$ is an explanation determined by a FA, while $v$ is a randomly generated explanation for the same input. For example the NC score of $u$ should be higher than $v$ when evaluating the diagnosticity of using NC as the faithfulness metric. More formally, diagnosticity $D_\varepsilon(F)$ is computed as follows:\footnote{For a proof of Eq.~\ref{eq:diagnosticity}, refer to \citet{chan-etal-2022-comparative}.}

\begin{equation}
    \begin{aligned}
    \label{eq:diagnosticity}
    {\displaystyle D_\varepsilon(F) \approx \frac{1}{|Z_\varepsilon|} \sum\limits_{(u,v)\in Z_\varepsilon} \mathbb{1}(u \succ _F v)}
    \end{aligned}
\end{equation}

\noindent where $F$ is a faithfulness metric, $Z_\varepsilon$ is a set of explanation pairs, also called $\varepsilon \text{-faithfulness}$ golden set, $0 \leq \varepsilon \leq 1$.
$\mathbb{1 \cdot}$ is the indicator function which takes a value 1 when the input statement is true and a value 0 when it is false. 

\citet{chan-etal-2022-comparative} randomly sample a subset of explanation pairs $(u, v)$ for each dataset and also randomly sample a FA for each pair. In our experiments, we do not sample but we  consider all the possible combinations of data points and FAs across datasets.

\section{Results}

\subsection{Diagnosticity of Faithfulness Metrics}

We compare the diagnosticity of faithfulness metrics introduced in Section \ref{sec:faithfulness_metrics}. Tables \ref{tab:D_avgon_DATA} and \ref{tab:D_avgon_FA}  show average diagnosticity scores across FAs and tasks, respectively. See App. \ref{app:faithfulness_results} for individual results for each faithfulness metric, FA and dataset.

In general, we observe that Soft-NC and Soft-NS achieve significantly higher diagnosticity scores (Wilcoxon Rank Sum, $p < .01$) than NC and NS across FAs and datasets. The average diagnosticity of Soft-NC is 0.529 compared to 0.394 of NC while the diagnosticity of Soft-NS is 0.462 compared to NS (0.349). 
Our faithfulness metrics outperform NC and NS in 16 out of 18 cases, with the exception of Soft-NC on AG and Soft-NS on M.RC.

In Table \ref{tab:D_avgon_DATA}, we note that both NC and Soft-NC consistently outperform Soft-NS and NS, which corroborates findings by \citet{chan-etal-2022-comparative}. We also see that using different FAs result into different diagnosticity scores. For example, diagnosticity ranges from 0.514 to .561 for Soft-NC while Soft-NS ranges from .441 to .480. We also observe similar behavior for NC and NS confirming results from \citet{atanasova-etal-2020-diagnostic}. Furthermore, we surprisingly see that various faithfulness metrics disagree on the rankings of FAs. For example DL is the most faithful FA measured by Soft-NC (.561) while NC ranks it as one of the least faithful (.372). However, Soft-NC and Soft-NS appear to be more robust by having less variance.

In Table \ref{tab:D_avgon_FA}, we observe that the diagnosticity of all four faithfulness metrics is more sensitive across tasks than FAs (i.e. wider range and higher variance).
Also, we notice that in AG and M.RC, there is a trade-off between (Soft-)NS and (Soft-)NC. For example, on AG, Soft-NC is .649, the highest among all tasks but Soft-NS is the lowest. This result may be explained by the larger training sets of AG (102,000) and M.RC (24,029), compared to SST (6,920) and Ev.Inf (5,789) which might make the model more sensitive to the task-specific tokens.

\begin{table}[!t]
\resizebox{\columnwidth}{!}{%
\renewcommand*{\arraystretch}{1.3}
\begin{tabular}{@{}lcccccc@{}}
\toprule
 & \multicolumn{1}{l}{$\alpha$} & \multicolumn{1}{l}{\begin{tabular}[c]{@{}l@{}}$\alpha\nabla\alpha$\end{tabular}} & \multicolumn{1}{l}{\begin{tabular}[c]{@{}l@{}}$x\nabla x $\end{tabular}} & \multicolumn{1}{l}{IG} & \multicolumn{1}{l}{DL} & \multicolumn{1}{l}{Average} \\ \midrule

NC & .404 & .405 & .358 & .428 & .372 & .394 (.025) \\
Soft-NC & \textbf{.525} & \textbf{.514} & \textbf{.526} & \textbf{.516} & \textbf{.561} & \textbf{.529}$^*$ (.017) \\ \cmidrule{2-7} 
NS & .400 & .383 & .300 & .368 & .294 & .349 (.044) \\
Soft-NS & \textbf{.479} & \textbf{.480} & \textbf{.444} & \textbf{.467} & \textbf{.441} & \textbf{.462}$^*$ (.017) \\
\bottomrule
\end{tabular}%
}
\caption{Diagnosticity of soft normalized comprehensiveness (Soft-NC) and sufficiency (Soft-NS) compared to AOPC (hard) normalized comprehensiveness (NC) and sufficiency (NS) across FAs. $^*$ denotes a significant difference compared to its counterpart on the same FA, $p < .01$. 
}
\label{tab:D_avgon_DATA}
\end{table}

\subsection{Qualitative Analysis}\label{sec:quali}
We further conduct a qualitative analysis to shed light on the behavior of faithfulness metrics for different explanation pairs consisting of real and random attribution scores. Table \ref{tab:quali} shows three examples from Ev.Inf, SST and AG respectively.

\paragraph{Repetitions in rationales affect faithfulness:}
Examining Example 1 (i.e. a biomedical abstract from Ev.Inf), we observe that the rationale (top 20\% most important tokens) identified by DL contains repetitions of specific tokens, e.g. ``aliskiren'', ``from'', ``in''. On one hand, ``aliskiren'' (i.e. a drug for treating high blood pressure) is the main subject of the biomedical abstract and have been correctly identified by DL. 
On the other hand, we observe that many of these repeated tokens might not be very informative (e.g. many of them are stop words), however they have been selected as part of the rationale. This might happen due to their proximity to other informative tokens such as ``aliskiren'' due to the information mixing happening because of the contextualized transformer encoder~\cite{tutek-snajder-2020-staying}.

\begin{table}[!t]
\resizebox{\columnwidth}{!}{%
\renewcommand*{\arraystretch}{1.3}
\begin{tabular}{lccccc}
\hline
 & SST & Ev.Inf & AG & M.RC & Average \\ \midrule
NC & .409 & .315 & .416 & \textbf{.434} & .394 (.046)\\
Soft-NC & \textbf{.431} & \textbf{.628}$^*$ & \textbf{.649}$^*$ & .406$^*$ & \textbf{.529}$^*$ (.111)  \\ \cmidrule{2-6} 

NS & .384 & .344 & \textbf{.385} & .282 & .349 (.042) \\
Soft-NS & \textbf{.467} & \textbf{.560}$^*$ & .294 & \textbf{.527}$^*$ & \textbf{.462}$^*$ (.102) \\  
\bottomrule
\end{tabular}%
}
\caption{Diagnosticity of faithfulness metrics across tasks. $^*$ denotes a significant difference compared to its counterpart on the same task, $p < .01$.}
\label{tab:D_avgon_FA}
\end{table}

We also notice that the random attribution baseline (Rand) selects a more diverse set of tokens that appear to have no connection between each other as expected. The random rationale also contains a smaller proportion of token repetitions. These may be the reasons why the random rationales may, in some cases, provide better information compared to the rationales selected by DL (or other FAs), leading to lower diagnosticity. Furthermore, NC between DL (.813) and Rand (.853) is very close (similar for NS) which indicates similar changes to predictive likelihood  when retaining or removing rationales by DL and Rand. However, this may misleadingly suggest a similar model reasoning on the two rationales. We observe similar patterns using other FAs. Incorporating the FA importance scores in the input embeddings helps Soft-NC and Soft-S to mitigate the impact of issues above as they use all tokens during the evaluation.

\begin{table*}[!t]
\scriptsize
\resizebox{\textwidth}{!}{%
\renewcommand*{\arraystretch}{1.2}
\begin{tabular}{|c|c|c|l|l|}
\hline
 & {\bf Text} & {\bf FA} & {\bf Metric} & {\bf Faith.} \\ \hline
 
\multirow{8}{*}{1} & \multirow{8}{*}{\parbox{9cm}{TITLE: Long-term effects of \colorbox{red!50}{aliskiren} \colorbox{pink}{on} blood pressure and the renin angiotensin - aldosterone system \colorbox{pink}{in} \colorbox{gray!10}{hypertensive} hemodialysis patients.
\colorbox{gray!22}{ABSTRACT}.OBJECTIVE: The long-term \colorbox{gray!25}{effects} of \colorbox{red!50}{aliskiren} \colorbox{pink}{in} hypertensive hemodialysis patients remain \colorbox{pink}{to} be elucidated. 
ABSTRACT.DESIGN: \colorbox{pink!42}{In} this \colorbox{gray!33}{post} hoc analysis\colorbox{gray!44}{,} we followed up 25 hypertensive hemodialysis patients who \colorbox{gray!62}{completed} 8-week \colorbox{red!50}{aliskiren} treatment in a \colorbox{gray!50}{previous} study for \colorbox{gray!30}{20} months \colorbox{pink!16}{to} investigate the blood pressure - lowering \colorbox{gray!45}{effect}.......
}} & \multirow{4}{*}{DL} 

 & NC & .813\\ 
 &  &  & Soft-NC & .984 \\
 &  &  & NS & .159\\ 
 &  &  & Soff-NS & .904 \\ 
 \cline{3-5} 
 & & \multirow{4}{*}{Rand}   
 & NC & .853 \\ 
 &  &  & Soft-NC & .351 \\
 &  &  & NS & .116 \\
 &  &  & Soff-NS & .055 \\ 

\hline
\hline 
\multirow{8}{*}{2} & \multirow{8}{*}{\parbox{9cm}{\colorbox{gray!25}{by} the \colorbox{gray!50}{end} i was \colorbox{gray!49}{looking} for something \colorbox{red!45}{hard} with which \colorbox{red!25}{to} bludgeon myself \colorbox{red!40}{unconscious}
}} & \multirow{4}{*}{$x\nabla x$} 
 & NC & .131 \\ 
&  &  & Soft-NC & .339 \\
&  &  & NS & .743 \\
&  &  & Soff-NS & .975 \\ \cline{3-5} 
  & & \multirow{4}{*}{Rand} 
 & NC & .097 \\
 &  &  & Soft-NC & .101 \\
 &  &  & NS & .787 \\
 &  &  & Soff-NS & .557 \\ 

\hline 
\hline
\multirow{8}{*}{3} & \multirow{8}{*}{\parbox{9cm}{\colorbox{red!25}{ATHENS}, \colorbox{red!25}{Greece} - Right now, \colorbox{red!33}{the} Americans are n't just a \colorbox{red!11}{Dream} \colorbox{gray!19}{Team} - they\colorbox{red!58}{'}re \colorbox{gray!24}{more} like \colorbox{red!40}{the} Perfect Team. Lisa Fernandez pitched a three - hitter Sunday and \colorbox{gray!19}{Crystl} \colorbox{red!46}{Bustos} drove \colorbox{gray}{in} two runs \colorbox{gray!55}{as} \colorbox{red!40}{the} Americans rolled to their eighth shutout \colorbox{gray!55}{in} \colorbox{gray!57}{eight} days\colorbox{red!58}{,} 5-0 over Australia , \colorbox{gray!66}{putting} them into \colorbox{red!40}{the} \colorbox{gray!15}{gold} medal \colorbox{gray!77}{game} ...}} & \multirow{4}{*}{IG} 
 & NC &  .186 \\
&  &  & Soft-NC & .997 \\
&  &  & NS & .016 \\
&  &  & Soff-NS & .962 \\ \cline{3-5} 
 
 & & \multirow{4}{*}{Rand} 
 & NC &  .194\\ 
 &  &  & Soft-NC & .003 \\
 &  &  & NS & .020 \\\
 &  &  & Suff-NS & .315 \\ 
 
 \hline
\end{tabular}%
}
\caption{Examples of inputs with their rationales (when taking the top 20\% important tokens) and their different faithfulness metrics scores. Highlighted tokens are the rationales by a given \colorbox{red!50}{FA} and the \colorbox{gray!55}{random} baseline. The tints indicate their importance scores, the lighter the less important.
The three examples are from Ev.Inf, SST and AG, respectively.
}
\label{tab:quali}
\end{table*}

\paragraph{Evenly distributed FA scores affect NC and NS:} 
We also notice that for some inputs, the token importance assigned by FAs is very close to each other as demonstrated in Example 3, i.e. a news article from AG. The evenly distributed importance scores lead to similar low NC and NS between the FA (IG) and the random baseline attribution. Considering that the FA scores and ranking truly reflect the model reasoning process (i.e. the model made this prediction by equally weighing all tokens), then the faithfulness measurements provided by NS and NC might be biased. 

We conjecture that this is likely to happen because these metrics entirely ignore the rest of the tokens even though these could represent a non-negligible percentage of the FA scores distribution. However, Soft-NC and Soft-NS take into account the whole FA distribution without removing or retaining any specific tokens, hence they do not suffer from this limitation.

\paragraph{Different part of speech preferences for tasks}
We find that FAs tend to favor different parts of speech for different tasks. 
In Example 1 where the task is to reason about the relationship between a given intervention and a comparator in the biomedical domain, FAs tend to select proper nouns (e.g. ``aliskiren'') and prepositions (e.g. ``on'', ``in'' and ``to'').
On the other hand, in Example 2 which shows a text from SST, FAs favor adjectives (e.g. ``unconscious'' and ``hard'') for the sentiment analysis task. In Example 3, we see that articles such as ``the'' and  proper nouns such as ``Greece'' and ``Bustos'' are selected.

\section{Impact of Rationale Length on Faithfulness and Diagnosticity}\label{sec:length}

Up to this point, we have only considered computing cumulative AOPC NC and NS by evaluating faithfulness scores at multiple rationale lengths together (see Section~\ref{sec:faithfulness_metrics}). Here, we explore how faithfulness and diagnosticity of NC and NS at individual rationale lengths compare to Soft-NC and Soft-NS. We note that both `soft' metrics do not take the rationale length into account. 

\subsection{Faithfulness}
\begin{figure}[!t] 
    \centering
    \includegraphics[width=0.494\columnwidth]{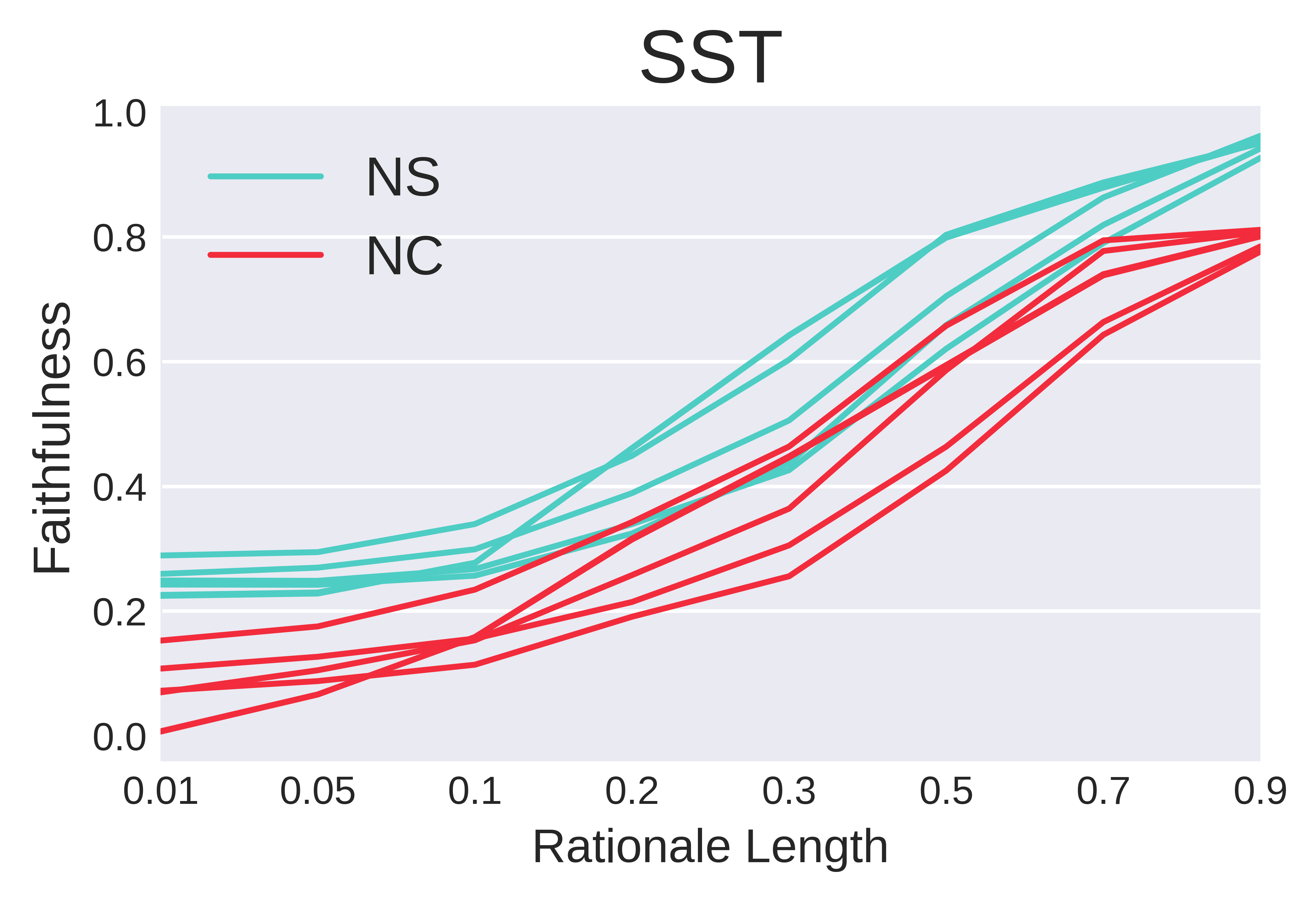}
    \includegraphics[width=0.494\columnwidth]{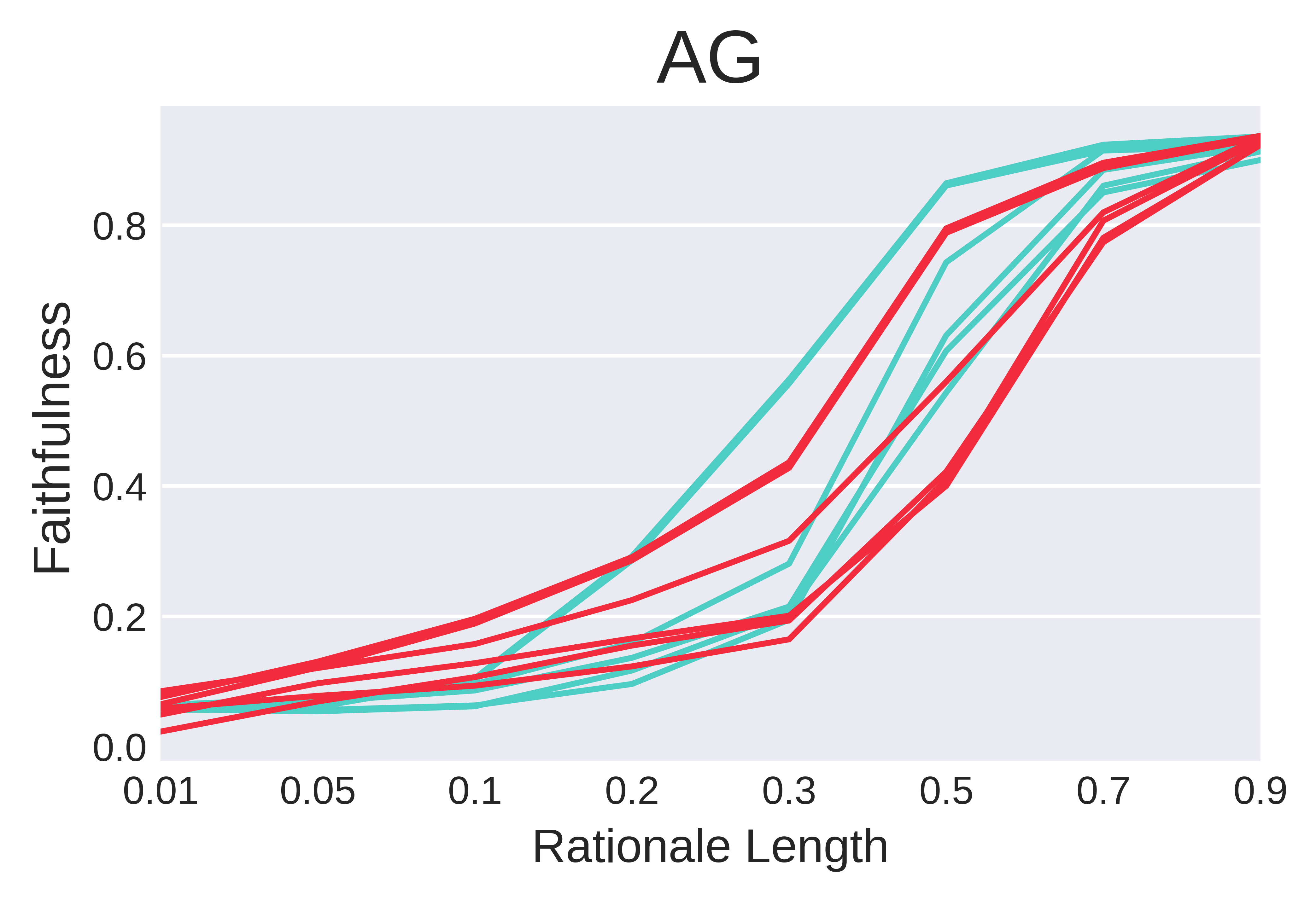}
    \includegraphics[width=0.494\columnwidth]{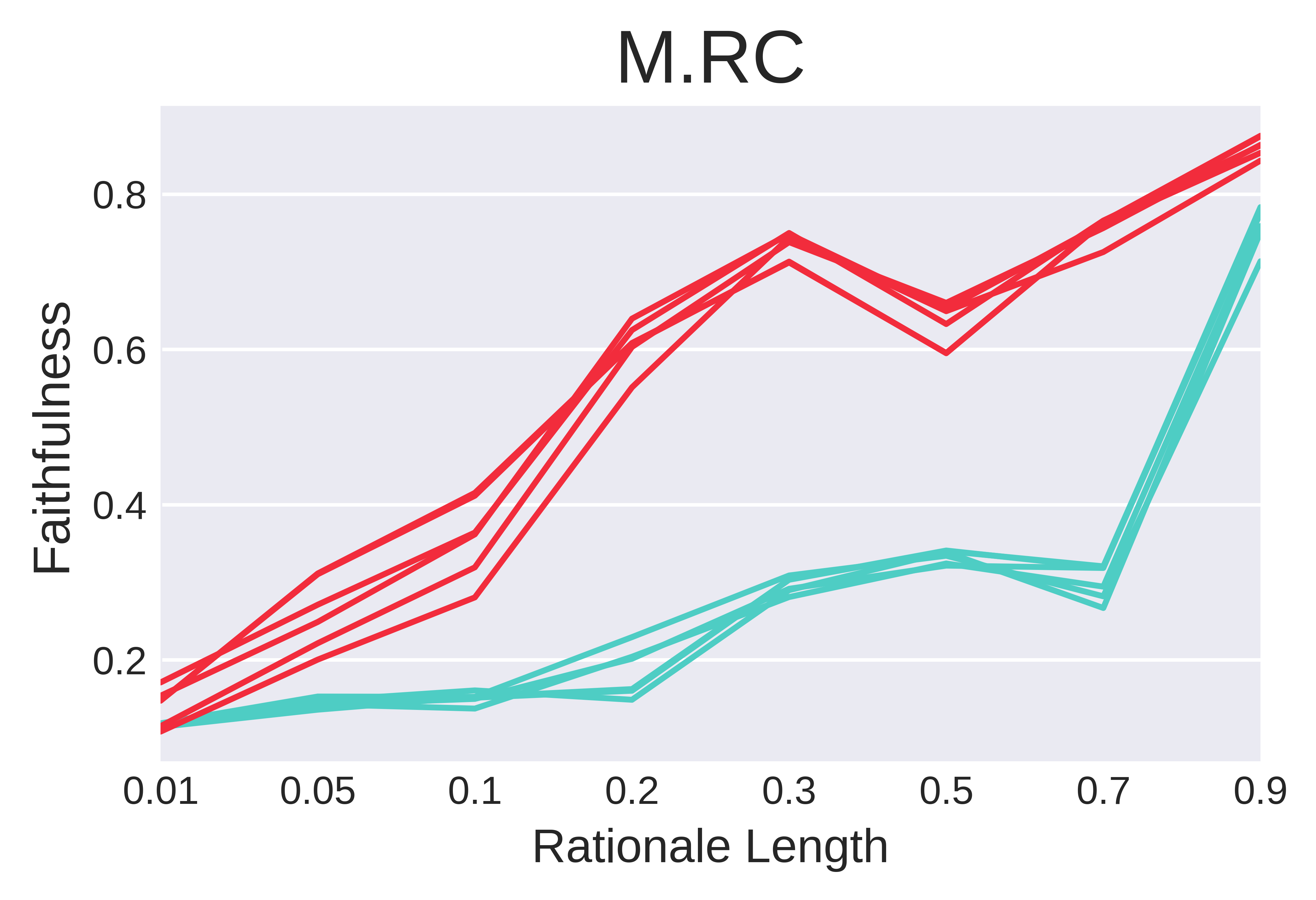}
    \includegraphics[width=0.494\columnwidth]{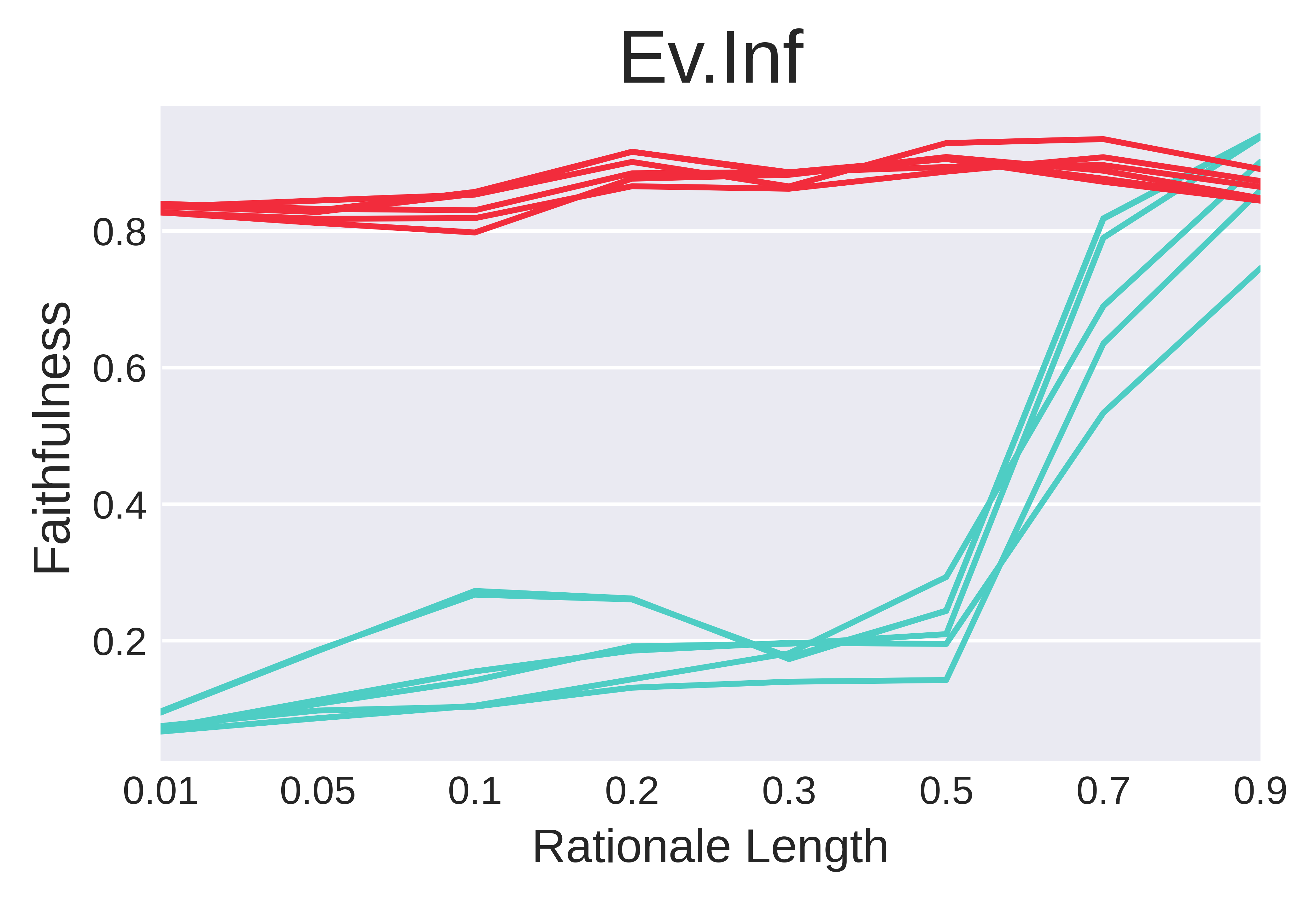}
    \caption{The impact of rationale length on normalized comprehensiveness (NC) and sufficiency (NS). Each line represents a FA.}\label{fig:len1}
\end{figure}

Figure \ref{fig:len1} shows the faithfulness scores of NC and NS at different rationale lengths for all FAs including random baseline attribution in each dataset.\footnote{For brevity, we do not highlight the different FAs as they follow similar patterns.}
We observe that the faithfulness scores of NC and NS follow an upward trend as the rationale length increases. This is somewhat expected because using information from an increasing number of tokens makes the rationale more similar to the original input.  

In AG and SST, NC and NS lines appear close by or overlap. One possible reason is that the input text in SST and AG is relatively short (average length of 18 and 36 respectively), possibly leading to higher contextualization across all tokens.
Therefore, removing or retaining more tokens results in a similar magnitude of changes in predictive likelihood.

In M.RC and Ev.Inf, two comprehension tasks that consist of longer inputs (average length is 305 and 365 respectively), we observe a different relationship between NC and NS. For instance, NC in Ev.Inf tends to be less impacted by the rationale length. This maybe due to the token repetitions in rationales discussed in Section \ref{sec:quali}. For example, when taking 2\% of the top-k tokens out, e.g. 6 out of 300 tokens, all the task-related tokens may have been removed already. 

\subsection{Diagnosticity}
\begin{figure}[!t] 
    \centering
    \includegraphics[width=0.494\columnwidth]{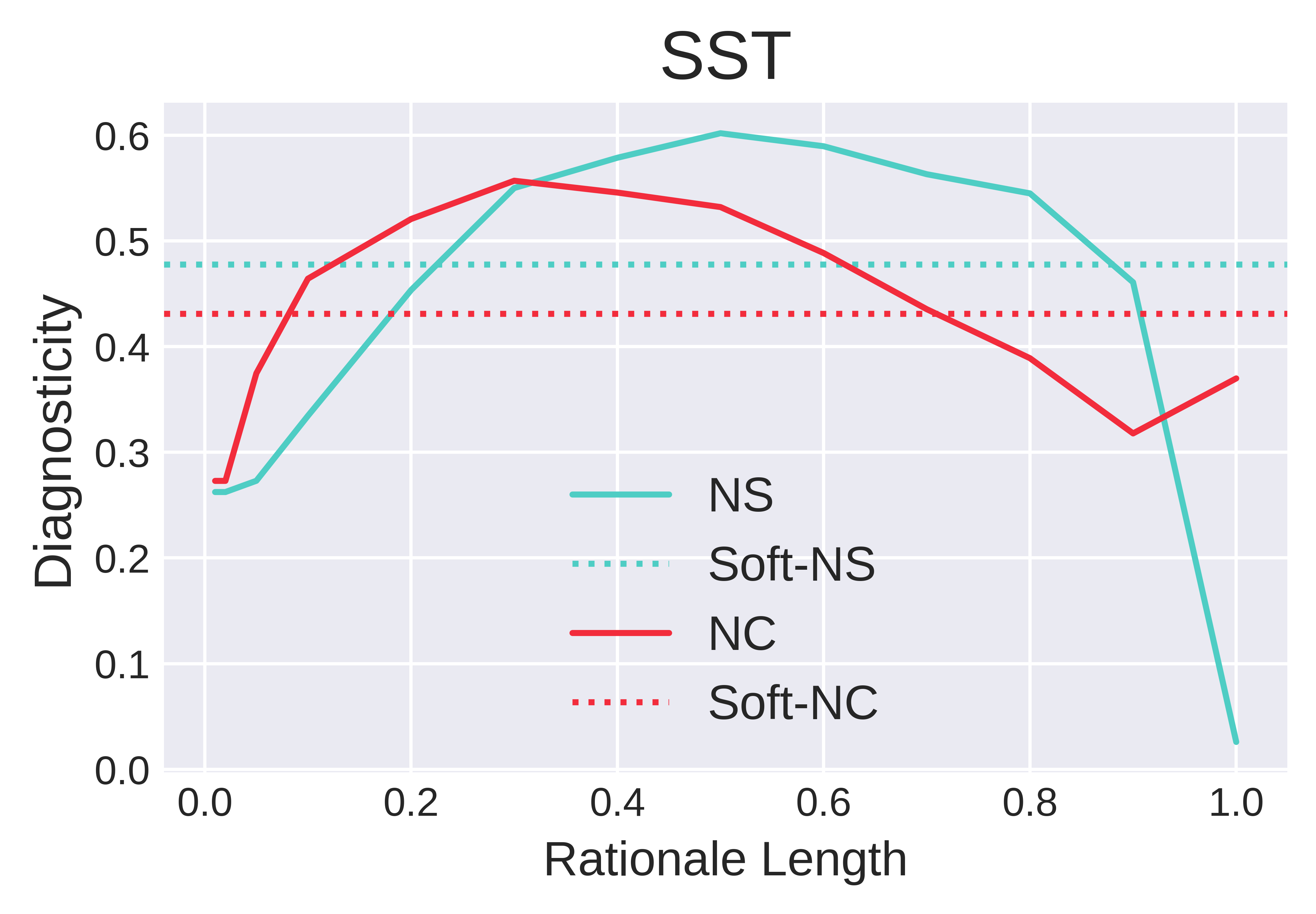}
    \includegraphics[width=0.494\columnwidth]{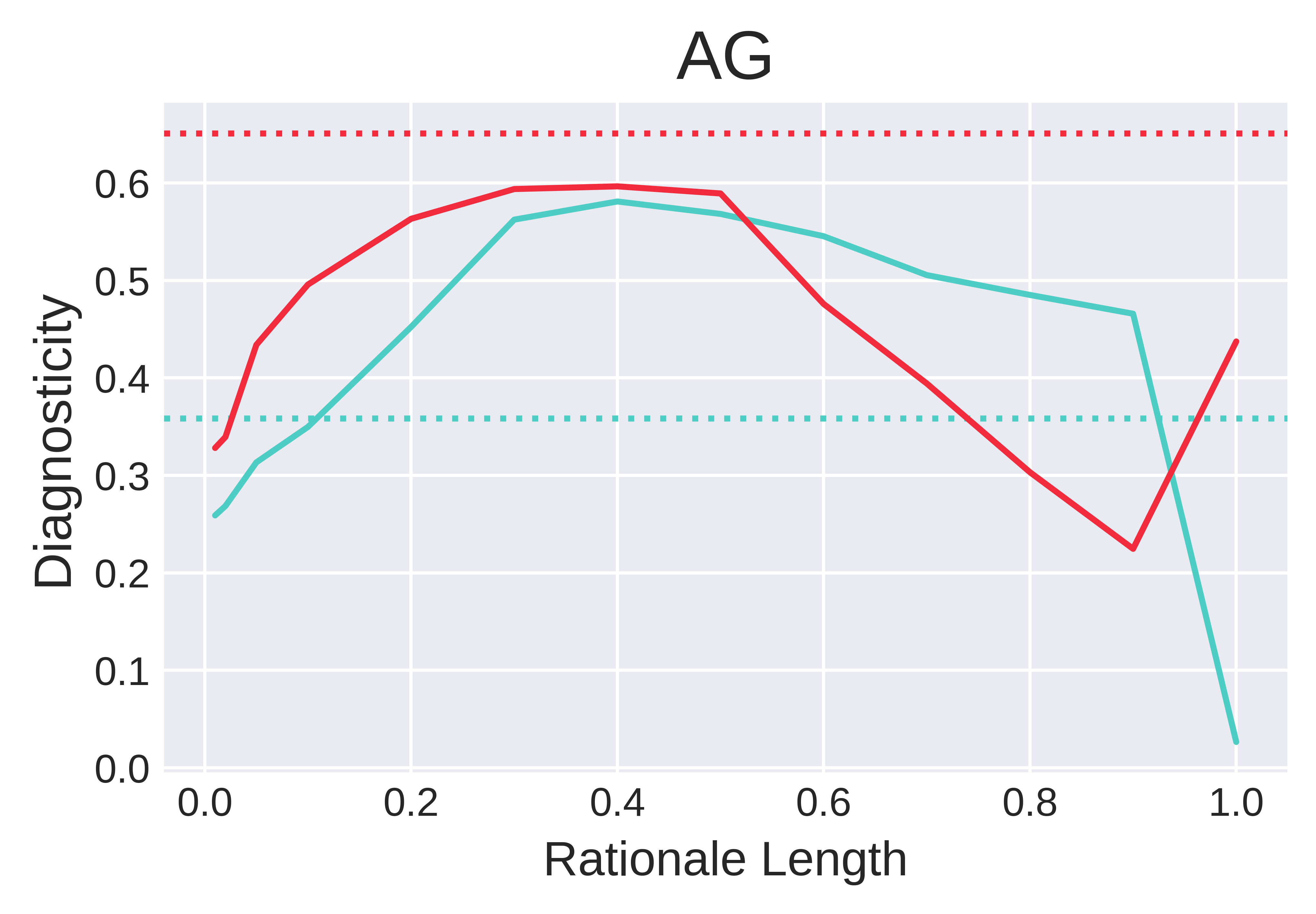}
    \includegraphics[width=0.494\columnwidth]{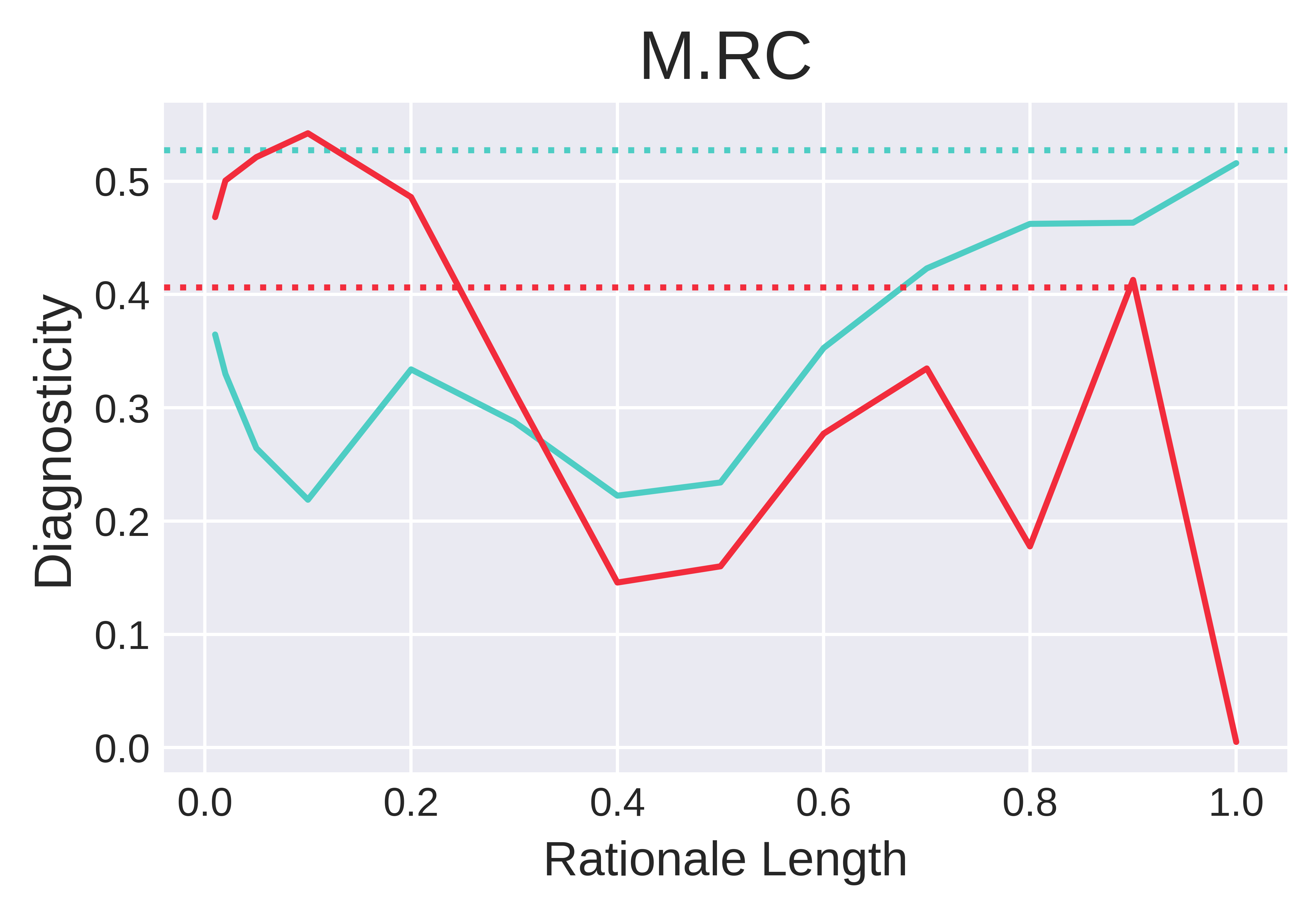}
    \includegraphics[width=0.494\columnwidth]{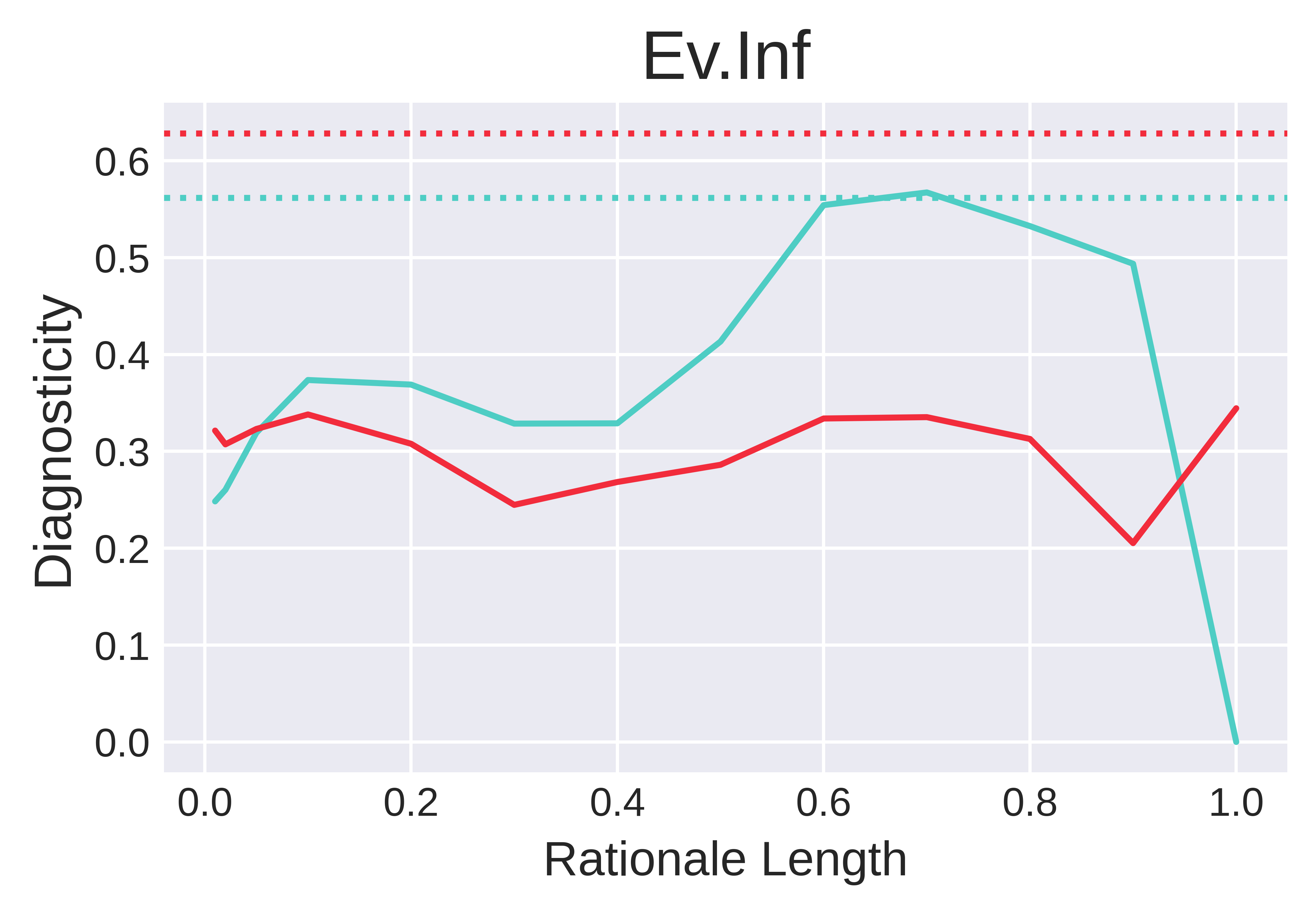}
    \caption{The impact of rationale length (shown in ratio) on Diagnosticity scores.}\label{fig:len_impact_on_dia}
\end{figure}

Figure \ref{fig:len_impact_on_dia} shows the diagnosticity scores of NS and NC on different rationale lengths (average across FAs) together with the diagnosticity of Soft-NC and Soft-NS. Overall in all datasets, we see that the diagnosticity of NC and NS does not monotonically increase as we expected. 
In SST and AG, the diagnosticity of NS and NC both initially increase and then decrease. This happens because after increasing to a certain rationale length, the random selected rationales (used in the diagnosticity metric) contain sufficient information making it hard for FAs to beat. 
In M.RC and Ev.Inf, Soft-NC and Soft-NS have higher diagnosticity than NC and NS. One possible reason is that the corrupted version of input could fall out-of-distribution, confusing the model. Our `soft' metrics mitigate this issue by taking all tokens into account.

Based on the observations on Figures \ref{fig:len1} and \ref{fig:len_impact_on_dia}, we conclude that it is hard to define an optimal rationale length for NC and NS which also has been demonstrated in previous work~\cite{chrysostomou2022flexible}.  
In general, we see that diagnosticity decreases along with longer rationale length for NC and NS. On the other hand, faithfulness measured by NC and NS increases for longer rationales (Figure \ref{fig:len1}). Therefore, this might be problematic for selecting optimal rationale length for NC and NS. For example, if we want to select an optimal rationale length for M.RC by looking at its relation to faithfulness, we might choose a length of 30\% over 20\% because it shows higher NC and NS. However, the diagnosticity of NC and NS is lower at 30\%, which means the higher NC and NS results to less trustful rationales. Our metrics bypass these issues because they focus on evaluating the FA scores and ranking as a whole considering all the input tokens. Soft-NC and Soft-NS do not require a pre-defined rationale length or evaluating faithfulness across different lengths.

We suggest that {\it it is more important to identify the most faithful FA given a model and task by taking into account all tokens rather than pre-defining a rationale of a specific length that ignores a fraction of the input tokens when evaluating faithfulness. The choice of how the FA importance scores will be presented (e.g. a top-k subset of the input tokens or all of them using a saliency map) should only serve practical purposes (e.g. better visualization, summarization of model rationales).}

\section{Conclusion}
In this work, we have proposed a new soft-perturbation approach for evaluating the faithfulness of input token importance assigned by FAs. Instead of perturbing the input by entirely removing or retaining tokens for measuring faithfulness, we incorporate the attribution importance by randomly masking parts of the token embeddings. Our soft-sufficiency and soft-comprehensiveness metrics are consistently more effective in capturing more faithful FAs across various NLP tasks. In the future, we plan to experiment with sequence labeling tasks. Exploring differences in faithfulness metrics across different languages is also an interesting avenue for future work.

\section*{Limitations}
This work focuses on binary and multi-class classification settings using data in English. Benchmarking faithfulness metrics in sequence labeling tasks as well as in multi-lingual settings should be explored in future work.

\section*{Acknowledgements}

ZZ and NA are supported by EPSRC grant EP/V055712/1, part of the European Commission CHIST-ERA programme, call 2019 XAI: Explainable Machine Learning-based Artificial Intelligence. This project made use of time on Tier 2 HPC facility JADE2, funded by EPSRC (EP/T022205/1). We thank Huiyin Xue for her invaluable feedback.

\bibliography{anthology,custom}
\bibliographystyle{acl_natbib}

\clearpage
\newpage
\appendix

\section{Model Hyperparameters}
\label{app:model_hyperparameters}

\begin{table}[!h]
\resizebox{\columnwidth}{!}{%
\begin{tabular}{@{}ccccc@{}}
\toprule
Dataset & Model & Batch Size & Learning Rate & Learning Rate (linear) \\ \midrule
SST & bert-base-uncased & 8 & 1e-5 & 1e-4 \\
AG & bert-base-uncased & 8 & 1e-5 & 1e-4 \\
Ev.Inf & scibert\_scivocab\_uncased & 4 & 1e-5 & 1e-4 \\
M.RC & roberta-base & 4 & 1e-5 & 1e-4 \\ \bottomrule
\end{tabular}%
}
\caption{Mode implementation details.}
\label{tab:hyper}
\end{table}

We use pre-trained models from the Huggingface library \citep{wolf-etal-2020-transformers}. We use the AdamW optimizer \citep{loshchilov2017decoupled} with an initial learning rate of $1e^{-5}$ for fine-tuning BERT. We fine-tune all models for 3 epochs using a linear scheduler, with 10\% of the data in the first epoch as warming up. We also use a grad-norm of 1.0. The model with the lowest loss on the development set is selected. All models are trained across 5 random seeds, and we report the average. Experiments are run on a single Nvidia Tesla V100 GPU. Table~\ref{tab:hyper} shows an overview of models and hyperparameters.

\section{Detailed Diagnosticity Results}
\label{app:faithfulness_results}

\begin{table}[!h]
\resizebox{\columnwidth}{!}{%
\begin{tabular}{@{}llcccc@{}}
\toprule
\multicolumn{1}{c}{Dataset} & \multicolumn{1}{c}{Feature} & NS & Soft-NS & NC & Soft-NC \\ \midrule
SST & Attention & 0.406 & 0.496 & 0.349 & 0.407 \\
SST & Scaled attention & 0.387 & 0.509 & 0.352 & 0.396 \\
SST & Gradients & 0.324 & 0.495 & 0.394 & 0.394 \\
SST & IG & 0.437 & 0.489 & 0.535 & 0.395 \\
SST & Deeplift & 0.367 & 0.347 & 0.413 & 0.562 \\

Ev.Inf & Attention & 0.437 & 0.583 & 0.334 & 0.632 \\
Ev.Inf & Scaled attention & 0.448 & 0.576 & 0.329 & 0.624 \\
Ev.Inf & Gradients & 0.280 & 0.494 & 0.282 & 0.638 \\
Ev.Inf & IG & 0.294 & 0.564 & 0.298 & 0.615 \\
Ev.Inf & Deeplift & 0.263 & 0.582 & 0.331 & 0.633 \\

AG & Attention & 0.465 & 0.294 & 0.505 & 0.654 \\
AG & Scaled attention & 0.432 & 0.302 & 0.512 & 0.640 \\
AG & Gradients & 0.320 & 0.294 & 0.314 & 0.658 \\
AG & IG & 0.452 & 0.283 & 0.435 & 0.647 \\
AG & Deeplift & 0.256 & 0.296 & 0.315 & 0.648 \\

M.RC & Attention & 0.292 & 0.541 & 0.427 & 0.408 \\
M.RC & Scaled attention & 0.266 & 0.533 & 0.428 & 0.397 \\
M.RC & Gradients & 0.276 & 0.493 & 0.443 & 0.415 \\
M.RC & IG & 0.288 & 0.529 & 0.445 & 0.411 \\
M.RC & Deeplift & 0.290 & 0.538 & 0.428 & 0.400 \\ \bottomrule
\end{tabular}%
}
\caption{The diagnosticity of faithfulness metrics.}
\label{tab:faithfulness_results}
\end{table}

\section{Alternative implementations for soft perturbation}
\label{app:alternative_soft_perturbation}

\paragraph{Adding Gaussian noise}
We perturb the pre-trained word embeddings with standard Gaussian noise.
This Gaussian noise-based embedding perturbation is similar to the ``statistical noise'' used by \citet{zhang2018word} and \citet{lakshmi-narayan-etal-2019-exploration} for data augmentation. Specifically, we:
\begin{enumerate}
  \item Multiply the token embedding with the token importance score, adding Gaussian noise. The resulting embedding is $\gamma \lambda \odot \mathbf{x_i}$ in Equation \ref{eqa:alter1}, where $\mathbf{x_i}$ is the original input embedding and $\lambda$ is the FA scores (importance degree), $\gamma$ is the hyperparameters based on the FA scores. $\odot$ is element-wise multiplication. As demonstrated by \citet{lakshmi-narayan-etal-2019-exploration}, adding Gaussian noise to the embedding requires tuning the standard deviation. Similarly, we tune the standard deviation ${\sigma ^{2} \in}$ \{0.005, 0.01, 0.05, 0.1, 0.5, 1, 2\} for soft-comprehensiveness and soft-sufficiency separately. 
  \item Add the embedding $\gamma \lambda \odot \mathbf{x_i}$, to the token embedding ($\mathbf{x_i}$) to obtain a perturbed embedding ($\mathbf{x'_i}$).
  \begin{equation}\label{eqa:alter1}
    {\displaystyle \mathbf{x'_i} = \mathbf{x_i}+\gamma \lambda \odot \mathbf{x_i}, \gamma\sim {\mathcal {N}}(\mu ,\sigma ^{2})}
    \end{equation}
    
\end{enumerate}

An alternative way to add noise is to:
\begin{enumerate}
  \item Generate a noise embedding by multiplying the token embedding with Gaussian noise with standard deviation, $\sigma ^{2}$, associated with the importance score of the token. 
  The embedding $\gamma \odot \mathbf{x_i}$ in Equation \ref{eqa:alter2}, where $\mathbf{x_i}$ is the original input embedding and $\lambda$ is the importance score. 
  \item Add $\gamma \odot \mathbf{x_i}$, to the token embedding ($\mathbf{x_i}$) to get the perturbed embedding ($\mathbf{x'_i}$).
  \begin{equation}\label{eqa:alter2}
    {\displaystyle \mathbf{x'_i} = \mathbf{x_i}+\gamma \odot \mathbf{x_i}, \gamma\sim {\mathcal {N}}(\mu ,\sigma ^{2})}
    \end{equation}
\end{enumerate}

\paragraph{Continuous attention mask} We simply replace the binary-valued attention mask with a continuous-valued mask, where the continuous value is associated with the FA score for each token. The remaining part of the embeddings and the model remain the same.

\end{document}